# Deep Residual Network for Off-Resonance Artifact Correction with Application to Pediatric Body Magnetic Resonance Angiography with 3D Cones

David Y Zeng[1], Jamil Shaikh[2], Dwight G Nishimura[1], Shreyas S Vasanawala[2], Joseph Y Cheng[2]
[1]Electrical Engineering, Stanford University, Stanford, CA, USA
[2]Radiology, Stanford University, Stanford, CA, USA

**Abstract**

**Purpose:** Off-resonance artifact correction by deep-learning, to facilitate rapid pediatric body imaging with a scan time efficient 3D cones trajectory.

**Methods:** A residual convolutional neural network to correct off-resonance artifacts (Off-ResNet) was trained with a prospective study of 30 pediatric magnetic resonance angiography exams. Each exam acquired a short-readout scan (1.18 ms ± 0.38) and a long-readout scan (3.35 ms ± 0.74) at 3T. Short-readout scans, with longer scan times but negligible off-resonance blurring, were used as reference images and augmented with additional off-resonance for supervised training examples. Long-readout scans, with greater off-resonance artifacts but shorter scan time, were corrected by autofocus and Off-ResNet and compared to short-readout scans by normalized root-mean-square error (NRMSE), structural similarity index (SSIM), and peak signal-to-noise ratio (PSNR). Scans were also compared by scoring on eight anatomical features by two radiologists, using analysis of variance with post-hoc Tukey's test. Reader agreement was determined with intraclass correlation.

**Results:** Long-readout scans were on average 59.3% shorter than short-readout scans. Images from Off-ResNet had superior NRMSE, SSIM, and PSNR compared to uncorrected images across ±1kHz off-resonance ($P<0.01$). The proposed method had superior NRMSE over -677Hz to +1kHz and superior SSIM and PSNR over ±1kHz compared to autofocus ($P<0.01$). Radiologic scoring demonstrated that long-readout scans corrected with Off-ResNet were non-inferior to short-readout scans ($P<0.01$).

**Conclusion:** The proposed method can correct off-resonance artifacts from rapid long-readout 3D cones scans to a non-inferior image quality compared to diagnostically standard short-readout scans.

# Introduction

Magnetic resonance imaging (MRI) has the ability to provide critical information for pediatric angiography exams without ionizing radiation, as needed for computed tomography (1,2). However, pediatric MRI often involves sedation or general anesthesia (S/GA) to minimize bulk and respiratory motion to improve image quality, especially for uncooperative patients (3). To minimize the duration and intensity of S/GA and to potentially eliminate S/GA completely, rapid imaging is essential. Many techniques for acceleration via parallel imaging and compressed sensing have been developed and require advanced reconstruction algorithms (4–6).

Scan efficiency can also be achieved through sophisticated k-space sampling trajectories. One such trajectory is 3D cones (7), which is additionally motion-robust (8) and has intrinsic navigation for correcting artifacts from motion (9). However, the flexibility of data acquisition strategies has been limited by systemic errors such as signal phase error from off-resonance, which can be significant in the body (10,11). In the context of rapid imaging, a trade-off exists between increasing sampling efficiency with longer readouts and reducing image artifacts. Several existing techniques for off-resonance correction involve acquiring a field map but require additional scan time (12,13). Alternatively, automatic and semi-automatic techniques based upon single acquisitions are computationally time-consuming (14–16). Here, we investigate the use of deep learning to robustly reduce the impact of one major source of image artifacts to increase the flexibility in designing more efficient trajectories.

Deep learning can theoretically model phenomena such as off-resonance, and off-the-shelf hardware and software have been optimized for deep neural networks. A deep-learning approach can address both issues of scan and computational time for off-resonance correction. Thus, we present a deep learning, off-resonance artifact correction method for 3D cones trajectories that requires no additional scan time and is computationally efficient.

Our purpose was to develop a deep-learning technique to correct off-resonance artifacts in order to facilitate rapid pediatric angiography with a time-efficient 3D cones trajectory while maintaining image quality. We achieved the acceleration by extending the readout duration of the 3D cones trajectory and then correcting off-resonance artifacts with deep learning. We assessed the relative quality of the corrected images against images acquired with diagnostically standard readout durations.

# Theory

*Off-Resonance*

The off-resonance signal equation, ignoring relaxation, may be written as

$$s(t) = \int_r M(r) e^{-i2\pi r \cdot k_r(t)} e^{-i\Delta\omega_{0,r} t} dr.$$

The typical Fourier imaging equation is $\int_r M(r)e^{-i2\pi r \cdot k_r(t)} dr$. The off-resonance term, $e^{-i\Delta\omega_{0,r}t}$, multiplies the Fourier imaging term and thus has a convolutional effect in the image domain. The off-resonance term becomes more significant for longer readout durations, applying a phase error in k-space that manifests as blurring in the image. However, off-resonance, is spatially-varying so each voxel has a unique blurring kernel and thus the convolution is non-stationary. Example off-resonance blurring kernels are shown in Supporting Figure S1. for 3D cones trajectories of various readout durations. Importantly, although the point is blurred, the information remains local.

Framing off-resonance as a non-stationary deblurring problem, convolutional neural networks are well-suited to address off-resonance artifacts for 3D cones trajectories. First, the learned convolutional weights of the model may be interpreted as learning the corresponding deconvolutional weights of the forward off-resonance blurring and the non-linearity of the model may address the non-stationarity of the blurring by adaptively adjusting deconvolutional weights for each spatial location. Second, the necessary information exists and is local. The blurring locally distributes information and the model can be designed such that all the necessary information is within the neuronal receptive field, so that the model does not need to generate completely new information to guide the deblurring.

**Methods**

*A. Network Architecture*

To correct the off-resonance artifacts, we developed and trained a supervised, 3D convolutional residual network that we have termed Off-ResNet (Figure 1) (23). The input to the network was a 3D image with two channels corresponding to real and imaginary components. The first layer convolved the input to the residual layer size, followed by three residual layers with filter depths of 128. The output layer was the final image with two channels corresponding to real and imaginary components. We chose to design a relatively small network with less capacity because the off-resonance blurring kernels are low-rank (14,24). We wanted to promote the network to learn off-resonance correction while preventing overfitting, which would likely result in the memorization of anatomy.

All convolution layers were 3D convolutions with kernel size 5 x 5 x 5 followed by rectified linear unit (ReLU) activations. We chose a relatively large kernel because the network is shallow but the neuronal receptive field must still be large enough to encompass the off-resonance blurring kernel size.

We chose an $L_1$-loss based upon the results of Zhao et al. (25). Compared to $L_2$-loss, $L_1$-loss produces superior objective and perceptual results. We did not use a structural similarity index (SSIM) loss because it has been noted to cause changes of brightness or shifts of colors in natural images (28). The network was trained in TensorFlow, optimized with Adam with an initial learning rate of 0.0001 (26,27).

*B. Dataset and Training*

This was a HIPAA-compliant, prospective study with approval from the institutional review board; patient-informed assent and consent were given. We collected consecutive scans, with short and long readouts, in each exam for training data. Thirty pediatric magnetic resonance angiography (MRA) exams with ferumoxytol-enhancement (Feraheme, AMAG Pharmaceuticals, Cambridge, MA; 0.1 ml/kg) and light anesthesia were acquired on a 3 T scanner (Discovery MR750; GE Healthcare; Waukesha, WI), with spoiled gradient echo 3D cones trajectories (17,18) with pseudo-random ordering (19). Patient inclusion criteria was a clinical 3 T scan with ferumoxytol. Ferumoxytol injection was given approximately 15 minutes prior to the data acquisition, ensuring contrast had essentially stable tissue distribution during the scans. Ferumoxytol has an approximately 15 hour half-life in humans, enabling data collection under near-identical contrast conditions for both short and long readout scans (20,21). Scan parameters and subject demographics are shown in Supporting Table S1. All scans were reconstructed with gridding, and coil combined using ESPIRiT sensitivity maps with no motion correction (22). As observed in Ref. (19), the cones sampling with pseudo-random ordering resulted in minimal image artifacts from motion.

Short-readout scans were used as reference images. To create training input data, the k-space of the short-readout scans were re-gridded onto four 3D cones trajectories of increasing readout duration, for increasingly worse off-resonance artifacts. To further augment the data, simulated, zero-order, global off-resonance was then added to the k-space data and the augmented images were reconstructed. The simulated off-resonance was added at 101 frequencies between ±500 Hz. Supporting Figure S1. demonstrates the local blurring effects of various readout durations and off-resonances. The reference images were uncorrected for off-resonance. The corresponding long-readout images were used for qualitative validation. Light anesthesia limited bulk motion but inter-scan variation due to respiratory motion was still present, limiting the usefulness of long-readout images during training.

The network was fully convolutional so it could accept any size 3D input. We divided the 30 exams such that eight exams were augmented for training and 22 exams were used for testing. No patients were in both the training and testing sets. For training, each scan was divided into overlapping patches of 64 x 64 x 64 voxels, for a total of 847,000 patches, to further augment data and for fitting data onto GPU memory. This setup resulted in a training time of 32 days for eight epochs. For validation and testing, matrix size varied but a typical size was 420 x 420 x 120 voxels.

*C. Evaluation*

We used autofocus as our comparison technique for conventional correction. Autofocus is a leading technique for blind off-resonance correction without a field map (14). Autofocus exhaustively simulates candidate off-resonances and selects the frequency that minimizes a metric. The metric used in this study minimized the imaginary component of the image after removing low-frequency phase. However, autofocus requires long computational times that are impractical for clinical use. Furthermore, the optimization may fall into local minima, which can result in poor image quality.

For computational image quality metrics, we evaluated normalized root-mean-square error (NRMSE), SSIM, and peak signal-to-noise ratio (PSNR) as a function of off-resonance. Off-resonance performance was simulated over a range of ±1 kHz, twice the range of data augmentation used for training. Uncorrected short-readout and long-readout images and images corrected by Off-ResNet and autofocus were evaluated across the entire test set with the corresponding short-readout images as reference. For the long-readout images, the metric results were evaluated relatively, rather than absolutely, because of inter-scan variation between the long-readout and short-readout images. Differences in the metrics were evaluated by one-way analysis of variance (ANOVA) with post-hoc Tukey's test (29). $P < 0.01$ was considered significant for all experiments.

The metrics of the short-readout images were approached as a test-set evaluation to assess the performance with a known ground truth. However, the clinical application of the model was for correcting prospectively collected long-readout images. Thus, we performed additional qualitative assessments of Off-ResNet correcting the long-readout images.

For subjective image quality metrics, four images were simultaneously presented in a double-blind fashion to two board-certified radiologists (with 5 and 15 years of experience): uncorrected long-readout images, long-readout images with autofocus correction, long-readout images with Off-ResNet correction, and uncorrected short-readout images.

Image quality of the MRA exams was evaluated by assessing delineation of eight blood vessels. Vessels were evaluated on a 5-point scale: 5-sharp over entire course, 4-sharp over most of course, 3-can judge stenosis, 2-can judge course/patency, 1-cannot judge course/patency. Significance of difference in scores was determined by one-way ANOVA with post-hoc Tukey's test (29). Agreement between readers was calculated by intraclass correlation (ICC) (30).

**Results**

A short-readout scan from the test set was augmented with off-resonance and corrected with Off-ResNet and autofocus. Figure 2 shows that Off-ResNet can recover vessel sharpness in short-readout images across the range of ±500 Hz, within the range of the training data. In contrast, some distal vessel segments remain blurry after correction with autofocus. Validation loss closely tracked training loss, suggesting good generalization of the network (Supporting Figure S2. ).

The long-readout scans were on average 59.3% ± 10.4 (range, 47.0–81.7) shorter than the short-readout scans. Sample image results are shown in Figure 3. In Figure 3a-c the vessels are difficult to distinguish from the surrounding tissue. Autofocus is able to recover some vessels and Off-ResNet is able to recover sharper vessels with longer vessel segments. However, the reference image from the uncorrected short-readout image shows more small branches and longer vessel segments.

In Figure 3d, internal mammary vessels in the uncorrected long-readout image are severely blurred. In these cases, autofocus only partially recovers the vessels or incorrectly results in signal voids. In contrast, Off-ResNet has recovered some vessel sharpness and distinct vessels are visible. The short-readout image has the sharpest and most distinct vessels. For all datasets, Off-ResNet required less than a minute to compute the results on an Nvidia Titan Xp. For comparison, autofocus reconstruction required between 27 and 43 minutes.

Image quality metric results are shown in Figure 4. Off-ResNet had superior NRMSE, SSIM, and PSNR compared to autofocus ($P < 0.01$) over the entire range of ±1 kHz when correcting the short-readout image. When correcting the long-readout image, Off-ResNet had superior NRMSE compared to uncorrected images over the entire range of ±1 kHz ($P < 0.01$) and superior NRMSE compared to autofocus images over the range of -677 Hz to +1 kHz ($P < 0.01$). Off-ResNet also had superior SSIM and PSNR compared to both uncorrected and autofocus images over the entire range of ±1 kHz ($P < 0.01$).

Qualitatively, the images were assessed by reader scores and intraclass correlation (ICC) (Table 1). Importantly, Off-ResNet images were non-inferior to all images, including the uncorrected short-readout images, for all evaluated anatomical features. The uncorrected short-readout images were also non-inferior to all other images. Four ICC were excellent, one was good, and four were fair, as classified by Cicchetti (31).

**Discussion**

In this study we proposed a new deep-learning method for correcting off-resonance artifacts in scans with non-Cartesian trajectories. We tested the method on 3D cones trajectories for pediatric patient scans. The proposed method enabled scan time reduction by 59.3% with longer readout durations, while maintaining non-inferior image quality to diagnostically-standard scans, suggesting viability of accelerating 3D cones trajectories for pediatric body MRA.

Off-ResNet required less than a minute to correct each typical dataset. Off-ResNet also did not require additional scans or alterations to the k-space trajectory. Importantly, Off-ResNet requires less time to correct the image than the amount of scan time saved so the total exam time is still reduced and the workflow is unchanged. Fast correction is also important for promptly reviewing the images while the patient is still in the scanner to repeat the scan if image quality is poor.

Additionally, acceleration by increasing readout duration is independent of acceleration that could be obtained by parallel imaging and compressed sensing; further scan-time reduction can be achieved by combining these techniques. This acceleration could potentially allow for future work in considering a high-resolution 3D cones scan in a single breath hold.

From the example images, although Off-ResNet is non-inferior to the short-readout images, the short-readout images appear better for some features, such as in Figure 3d. The differences may be a result of more prominent $T_2^*$ effects as the readout is extended, or from inter-scan differences such as motion. Longer scans with periodic respiration under anesthesia will also have greater motion averaging. Correction for these artifacts are not accounted for when training the model. $T_2^*$ blurring may become the greatest limiting factor for readout duration, and thus acceleration, with this technique, exacerbated by the $T_2^*$-shortening effect of ferumoxytol (20). Acquiring a $T_2^*$ map and augmenting the training data with both off-resonance and $T_2^*$ may be a solution, as $T_2^*$ blurring is also a non-stationary, local artifact with 3D cones trajectories (32).

In addition to off-resonance and $T_2^*$ blurring, other sources of artifacts such as motion may also be framed as blurring with non-stationary kernels with localized information. Other trajectories with oversampled centers, such as 2D and 3D spiral-based trajectories (33,34), 2D and 3D radial-based trajectories (35,36), and FLORET (37) also have local blurring from off-resonance, suggesting that approaches similar to Off-ResNet may be suitable as well. Finally, although our study was focused on pediatric body imaging, Off-ResNet could correct for off-resonance artifacts in adult patients and in other anatomies.

Our study had several limitations. First, our study involved a small number of patients from a single institution. A larger cohort is necessary to confirm our findings. Additionally, a multi-center study is required to verify performance of the proposed method for robustness with scans with different imaging parameters and from different vendors. Second, we used a limited fraction of our available data to train. Future work would incorporate the testing data for training and collect new test data. Third, our primary metric for image quality was radiologist scores, which are subjective. Finally, we did not specifically evaluate the images for pathologies.

In conclusion, Off-ResNet is a deep-learning-based method for off-resonance artifact correction. It enables longer readouts and thus shorter scan times for pediatric MRA scans with 3D cones trajectories. Images corrected by Off-ResNet had non-inferior image quality as compared to images from scans that were significantly longer.

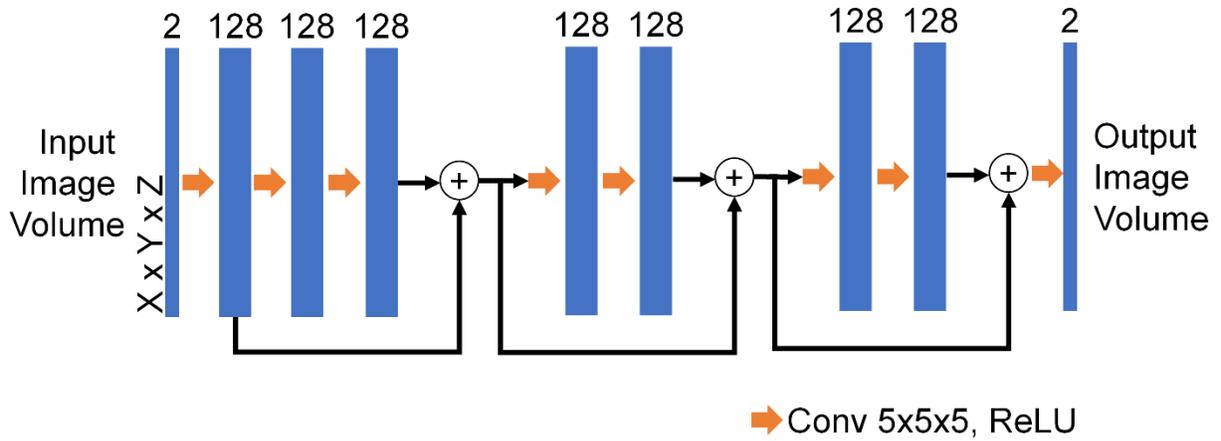

Figure 1. Off-ResNet Architecture. All 3D convolution kernels are 5 x 5 x 5 with rectified linear unit (ReLU) activations. The input and output are images with real and imaginary components as channels. The network is composed of an initial convolutional layer and three residual layers with two convolutional layers each.

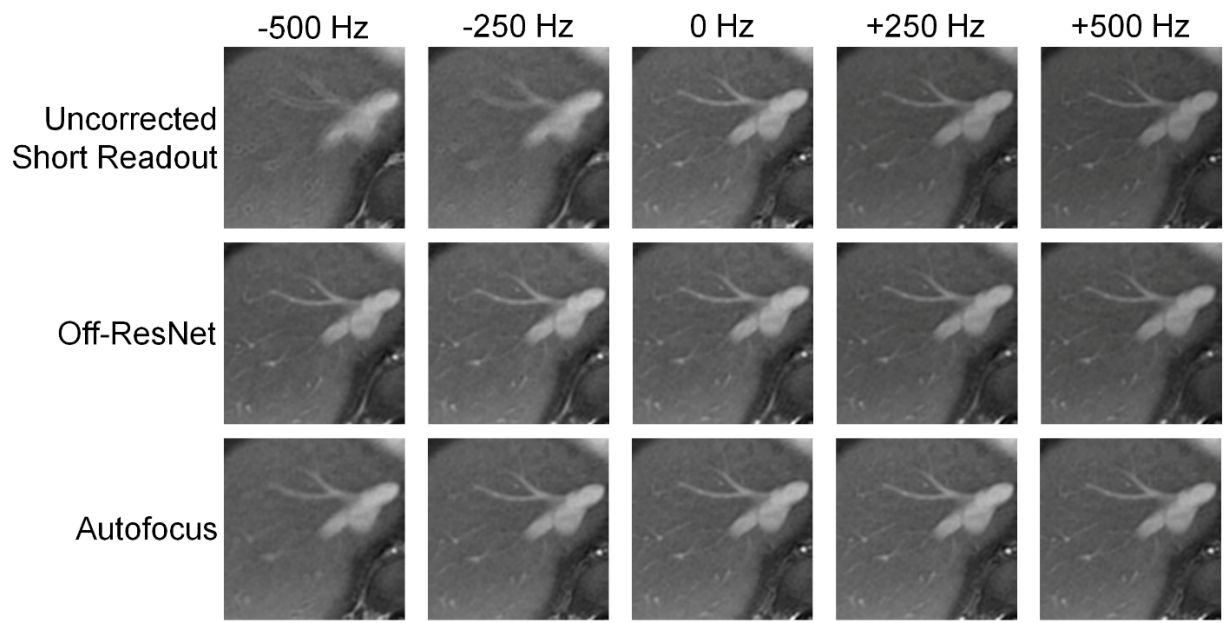

Figure 2. A short-readout scan from the test set, augmented with off-resonance and corrected with Off-ResNet and autofocus.

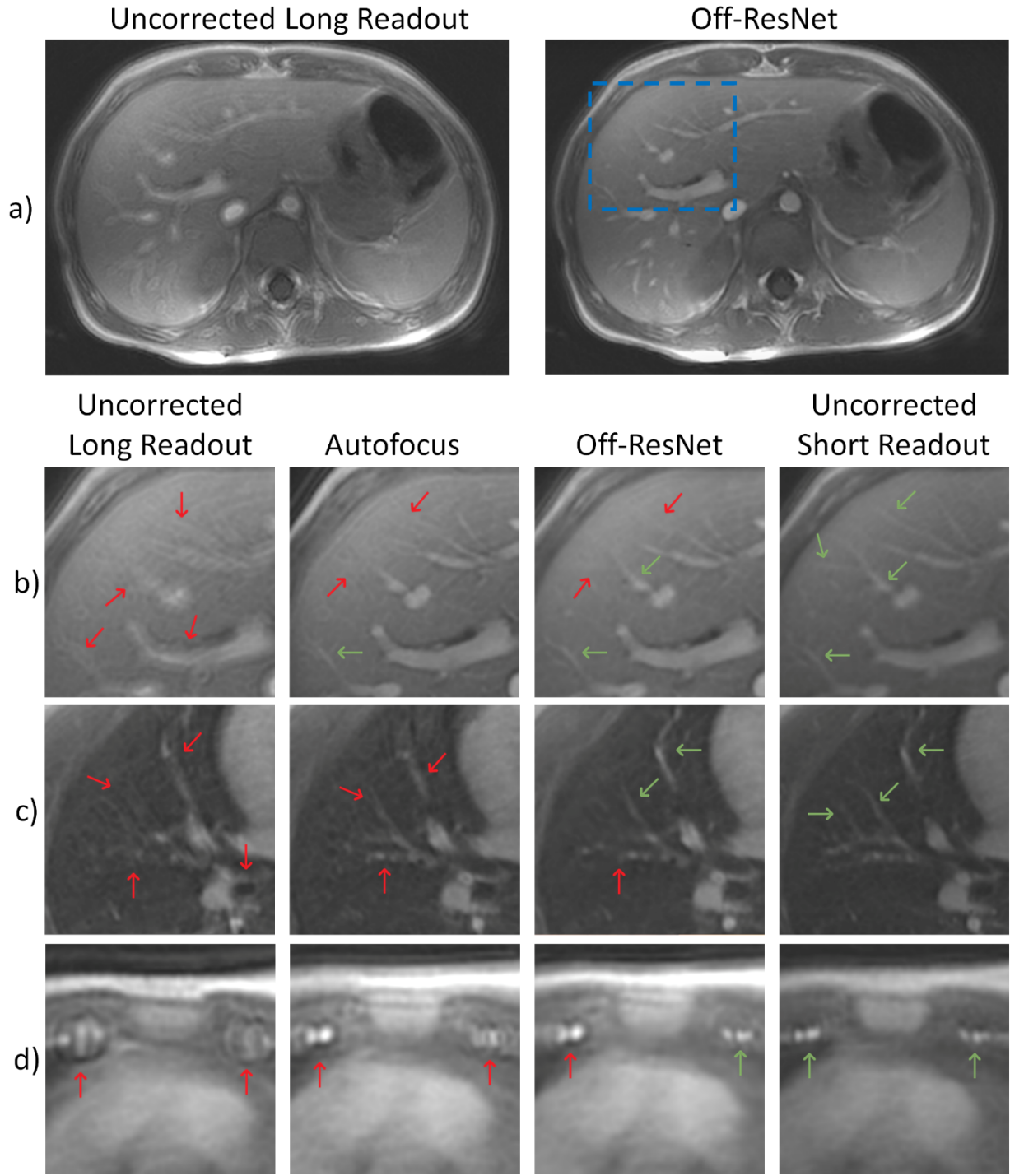

Figure 3. (a) An uncorrected long-readout and the same slice corrected by Off-ResNet. The blue box indicates the region in (b). (b-d) Sample images from the four categories compared by radiologists. Off-resonance blurring is most apparent in the loss of vessel sharpness (red arrows). Good vessel definition is highlighted by the green arrows. The (b) hepatic and portal veins, (c) subsegmental right pulmonary arteries, and (d) internal mammary arteries are shown.

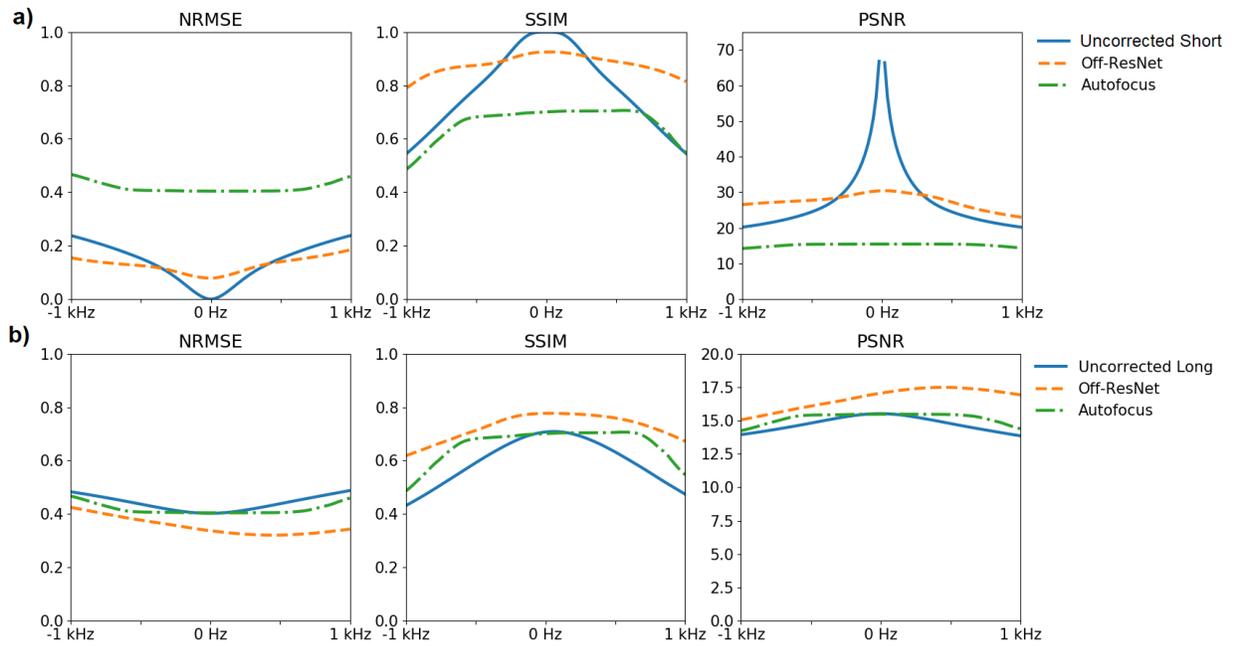

Figure 4. Off-resonance was added to (a) short-readout and (b) long-readout images and then corrected by Off-ResNet and autofocus. NRMSE, SSIM, and PSNR as a function of off-resonance for the uncorrected and corrected images are shown, with the on-resonance uncorrected short-readout image as reference.

Table 1. Mean Reader Scores

| Anatomical Feature | Uncorrected Long Readout | | Autofocus | | Off-ResNet | | Uncorrected Short Readout | | ICC (P < 0.01) |
|---|---|---|---|---|---|---|---|---|---|
| | Reader 1 | Reader 2 | Reader 1 | Reader 2 | Reader 1 | Reader 2 | Reader 1 | Reader 2 | |
| Right PA | 4.04 | 3.87 | 4.22 | 4.13 | 4.41 | 4.33 | 4.50 | 4.27 | 0.5198 |
| RLL PA | 3.68 | 2.80 | 3.95 | 3.47 | 4.14 | 3.73 [L] | 4.36 [L] | 4.00 [L] | 0.5614 |
| RLL Sup Segment PA | 2.68 | 2.07 | 3.41 | 3.07 | 3.73 [L] | 3.40 [L] | 3.77 [L] | 3.73 [L] | 0.6691 |
| Subsegmental RLL Sup Segment PAs | 2.04 | 1.13 | 2.50 | 1.93 | 3.41 [A,L] | 2.33 [L] | 3.68 [A,L] | 3.07 [A,L] | 0.7564 |
| Anterior Spinal Artery | 1.36 | 1.60 | 1.68 | 2.00 | 1.64 | 2.00 | 2.00 | 2.07 | 0.5784 |
| Right Internal Mammary Artery | 2.45 [A] | 2.13 | 1.18 | 1.40 | 2.86 [A] | 2.80 [A] | 3.41 [A,L] | 3.47 [A,L] | 0.8250 |
| Right Hepatic Vein | 2.36 | 2.20 | 3.91 [L] | 3.33 [L] | 3.91 [L] | 3.40 [L] | 4.18 [L] | 3.80 [L] | 0.8390 |
| Aorta at Diaphragm | 3.59 | 3.20 | 4.32 [L] | 4.27 [L] | 4.64 [L] | 4.20 [L] | 4.86 [L] | 4.40 [L] | 0.7523 |

[L] This method is superior to uncorrected long readout images (P < 0.01).
[A] This method is superior to autofocus images (P < 0.01).
Note: Rows with superscripts passed one-way analysis of variance significance with post-hoc Tukey's test (P < 0.01). PA = pulmonary artery, RLL = right lower lobe, Sup = superior, ICC = intraclass correlation.

**Supporting Methods**

To visualize the effects of the deep-learning correction, low-resolution off-resonance field maps of the corrected image were calculated. Simulated off-resonance between ±500 Hz was added to the corrected image and the simulated images were compared against the uncorrected image based on the autofocus metric. On a voxel-wise basis, the frequency of the simulated image with the smallest autofocus metric over the range of ±1 kHz was designated as the off-resonance of the voxel. For comparison, true field maps were calculated with double-TE methods ($TE_1$ = 0.032 ms, $TE_2$ = 1.3 ms). The true field map was also blurred with a Gaussian kernel for a low-resolution map to compare with the low-resolution Off-ResNet field map. With generative machine learning models, there is a concern for hallucinating structure and small differences in the field maps suggest lower probabilities of hallucinations from Off-ResNet.

We also tested for hallucinations by applying Off-ResNet four consecutive times to an uncorrected long-readout image, such that the output of the previous iteration would be fed into the model again. We hypothesized that if the network were hallucinating new structure, it would be independent of the underlying image structure. Thus, if the network were to hallucinate, repeated application of the network should generate new structure each time. Differences between applications were evaluated by NRMSE.

**Supporting Results**

To test the limit of Off-ResNet, the off-resonance was augmented even further with images shown in Supporting Figure S3.. Notably, the performance of Off-ResNet begins to diminish at off-resonances worse than -1000 Hz and +2000 Hz.

Supporting Figure S4. shows the original image and its corresponding $\Delta\omega_0$ maps. Supporting Figure S4.e shows the difference between the low-resolution true field map and the Off-ResNet field map. Importantly, the average absolute difference is 29.1 Hz, suggesting that Off-ResNet is not hallucinating new structures into the image.

In Supporting Figure S5. , the network is applied four times and after the first application, the subsequent applications have differences on the order of 0.1% the first application, again suggesting that the network is not hallucinating new structures.

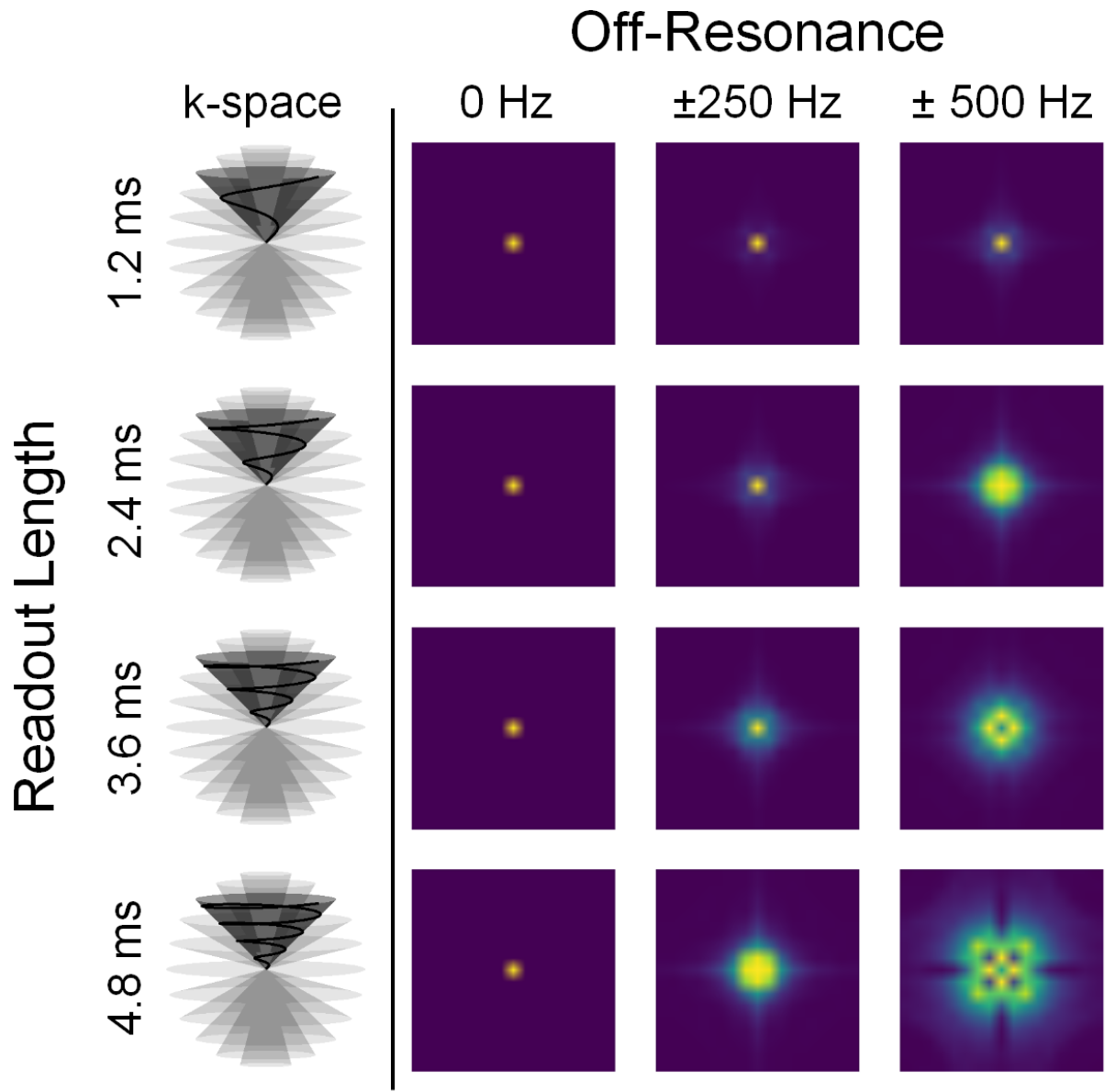

Supporting Figure S1. The left column shows the 3D cones trajectory in k-space versus readout time for one readout interleaf on an example cone shell. The right three columns show spatially-localized point spread functions of the 3D cones trajectory as a function of readout duration and off-resonance in the XY plane at Z = 0. Different off-resonances and trajectories have different point spread functions.

Supporting Table S1. Patient and Scan Parameters

| Characteristic | Short readout | Long readout |
|---|---|---|
| **Training Set (n = 8)** | | |
| Sex | | |
| Female | 4 | |
| Male | 4 | |
| Age* | 6.6 years ± 3.6 (2 – 12) | |
| | | |
| **Test Set (n = 22)** | | |
| Sex | | |
| Female | 12 | |
| Male | 10 | |
| Age* | 5.6 years ± 3.4 (0 – 11) | |
| | | |
| **Scan Parameters*** | | |
| Total scan time | 5.38 min ± 0.985 (3.05 - 7.99) | 2.19 min ± 0.565 (0.558 - 4.26) |
| Readout length | 1.18 ms ± 0.375 (0.944 - 2.43) | 3.35 ms ± 0.738 (2.43- 6.08) |
| Field of View XY | 30.0 cm ± 4.76 (20.0 – 36.0) | 30.1 cm ± 4.92 (20.0 – 38.0) |
| Field of View Z | 21.7 cm ± 3.65 (12.0 – 26.0) | 21.8 cm ± 3.75 (12.0 – 26.0) |
| Resolution XY | 0.744 mm ± 0.125 (0.481 – 1.00) | 0.746 mm ± 0.128 (0.481 – 1.00) |
| Resolution Z | 1.83 mm ± 0.292 (1.00 – 2.00) | 1.833 mm ± 0.292 (1.00 – 2.00) |
| Echo Time | 0.032 ms ± 0 (32 – 32) | 0.032 ms ± 0 (32 – 32) |
| Repetition Time | 5.17 ms ± 0.176 (5.00 – 5.60) | 7.44 ms ± 0.231 (6.80 – 8.00) |
| Flip | 15° ± 0 (15 – 15) | 15° ± 0 (15 – 15) |

* Data are means ± standard deviation (minimum – maximum)

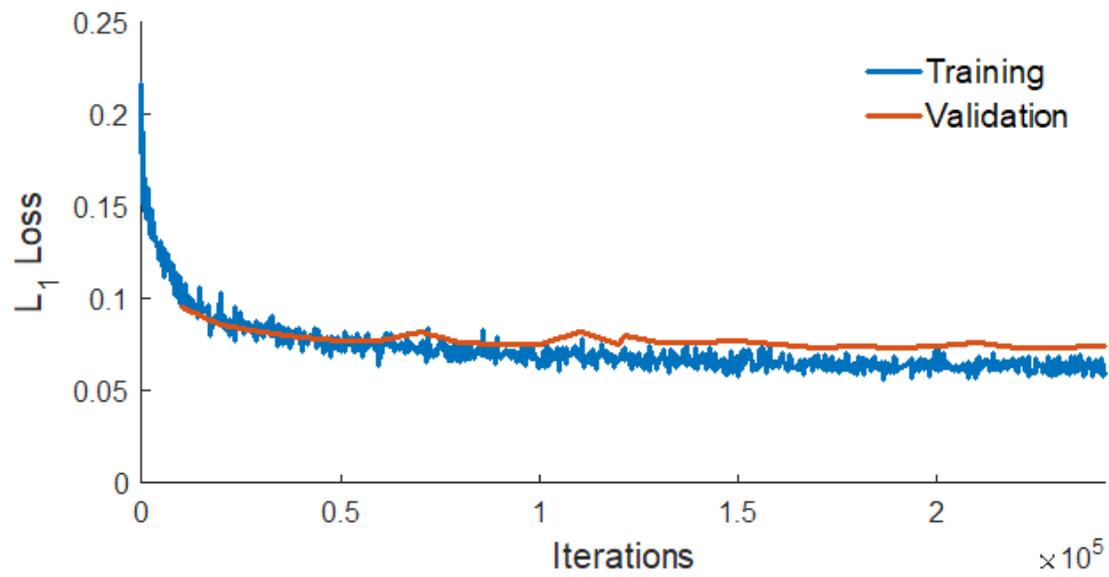
Supporting Figure S2. Loss graphs for training and validation sets over eight training epochs.

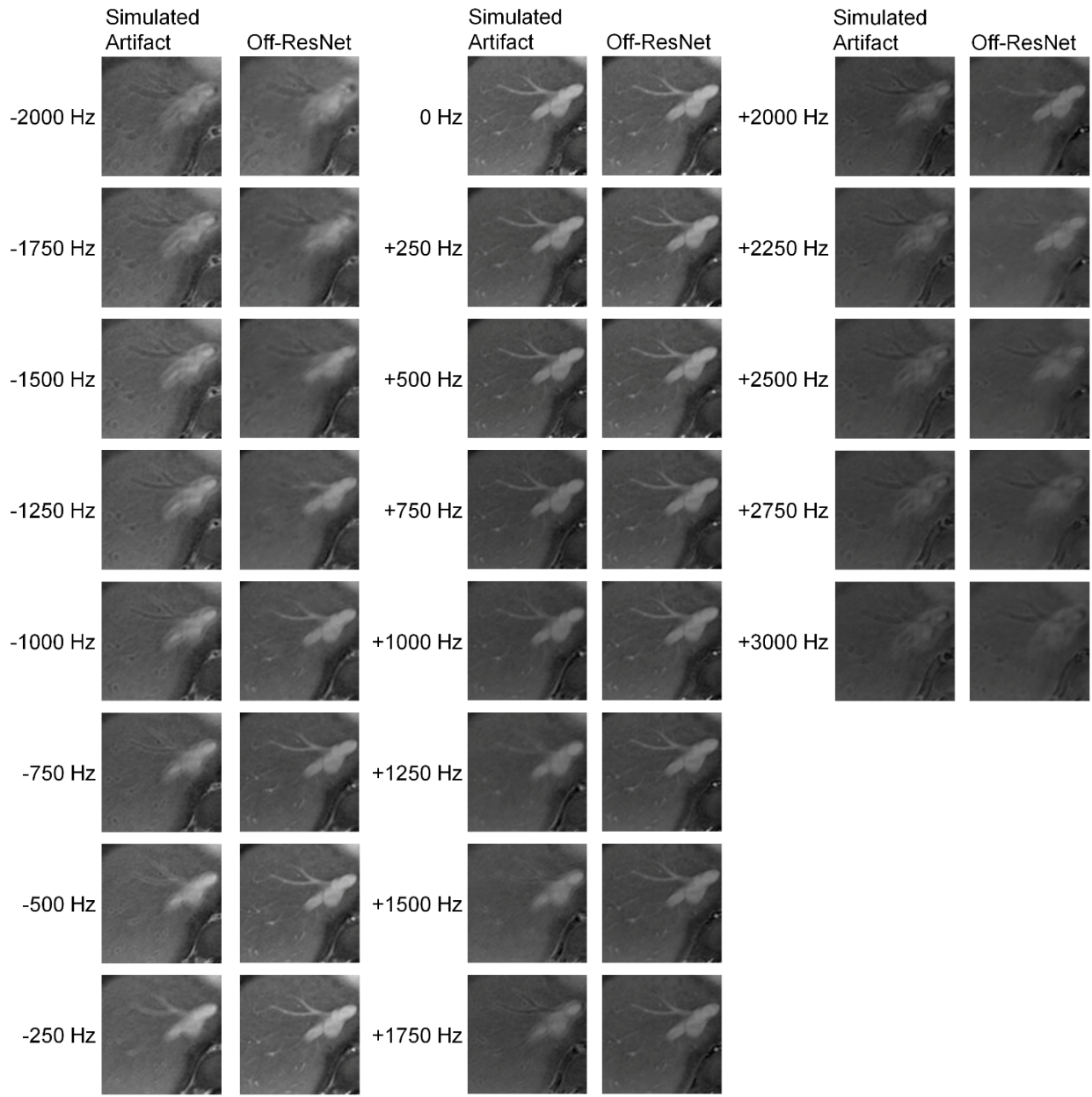
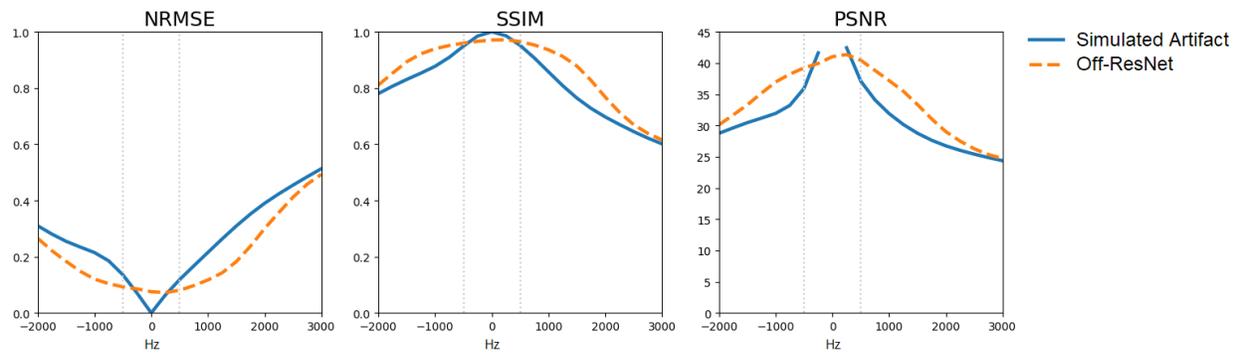

Supporting Figure S3. Extended off-resonance augmentation of short-readout scans. Off-ResNet performance begins to noticeably worsen at off-resonances worse than -1000 Hz and +2000 Hz.

Metrics for this dataset are shown. The network was trained at a range of -500 Hz to 500 Hz (gray vertical dotted lines).

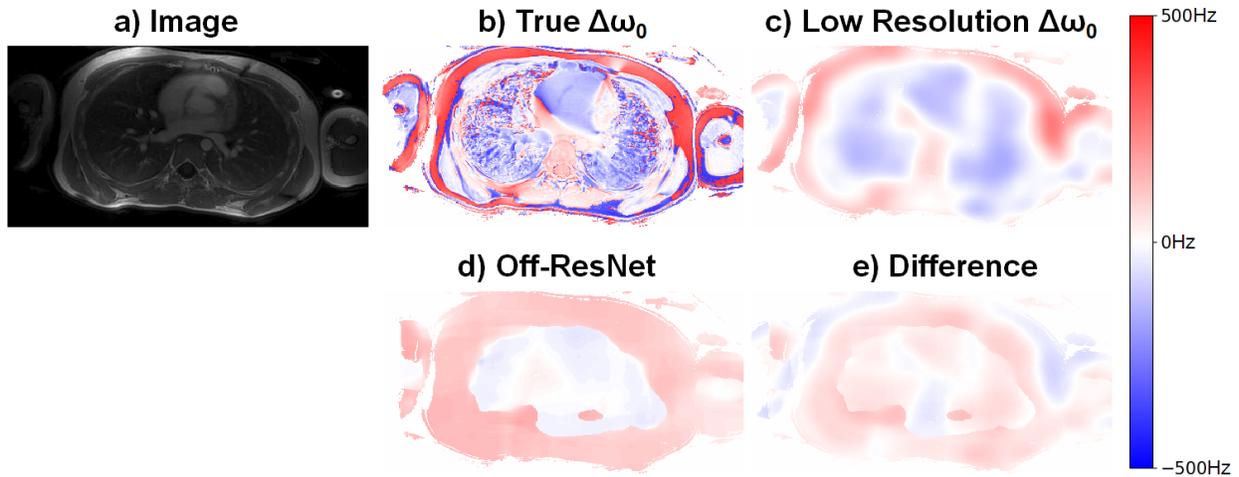

Supporting Figure S4. (a) The original image. (b) The true field map calculated by double-TE methods. (c) A low resolution field map of (b), blurred by a Gaussian kernel. (d) Field map of the output of Off-ResNet, calculated by simulating off-resonance in the output of Off-ResNet and finding the voxel-wise frequency which generates the image most similar to the original, uncorrected image, as determined by the autofocus metric. (e) The difference image between (c) and (d), demonstrating that the two field maps are similar, giving confidence that Off-ResNet is not hallucinating structures.

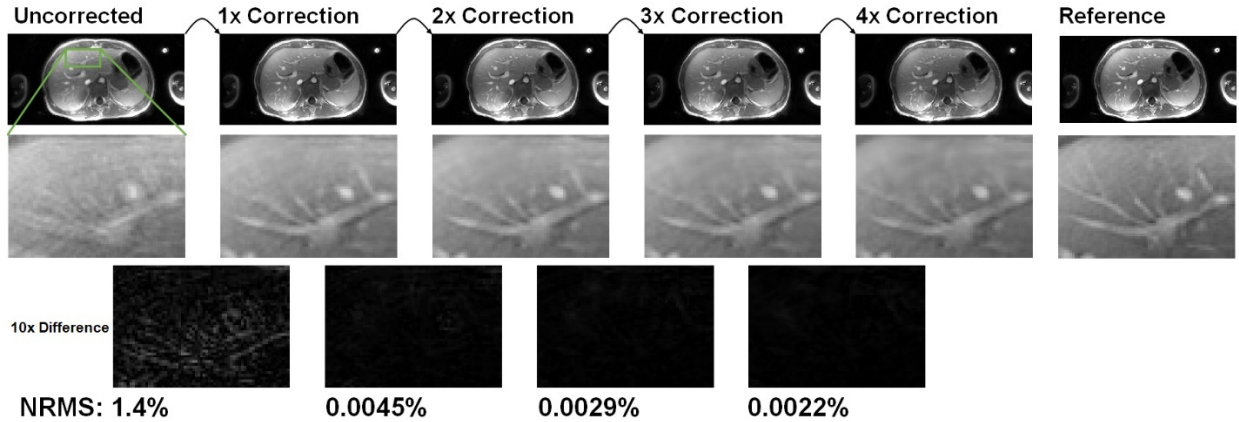

Supporting Figure S5. An uncorrected axial slice was corrected four consecutive times. The difference between each consecutive result is shown in the third row, increased by a factor of 10 for visibility. Normalized room-mean-square (NRMS) of the difference images were calculated and suggest that multiple applications of the network are approximately the same as applying the network once, suggesting that hallucinations from the network are unlikely.